\documentclass{article}

\PassOptionsToPackage{square, numbers}{natbib}
\usepackage[final]{nips_2016}
\usepackage{amsmath}
\usepackage[utf8]{inputenc} 
\usepackage[T1]{fontenc}    
\usepackage{hyperref}       
\usepackage{url}            
\usepackage{booktabs}       
\usepackage{amsfonts}       
\usepackage{nicefrac}       
\usepackage{microtype}      
\usepackage{graphicx}
\usepackage{algorithm} 
\usepackage{algpseudocode} 
\usepackage[font=small,labelfont=bf,center]{caption}

\hypersetup{colorlinks=true,urlcolor=blue,linkcolor=black,citecolor=black}
    
\title{FlapAI Bird: Training an Agent to Play Flappy Bird Using Reinforcement Learning Techniques}

\author{
  \and
  \textbf{Tai Vu}\\
  Department of Computer Science\\
  Stanford University\\
  \texttt{taivu@stanford.edu}
  \and
  \textbf{Leon Tran}\\
  Department of Computer Science\\
  Stanford University\\
  \texttt{leontk@stanford.edu}
}

\begin{document}

\maketitle

\begin{abstract}
    Reinforcement learning is one of the most popular approaches for automated game playing. This method allows an agent to estimate the expected utility of its state in order to make optimal actions in an unknown environment. We seek to apply reinforcement learning algorithms to the game Flappy Bird. We implement SARSA and Q-Learning with some modifications such as $\epsilon$-greedy policy, discretization and backward updates. We find that SARSA and Q-Learning outperform the baseline, regularly achieving scores of 1400+, with the highest in-game score of 2069.

\end{abstract}

\section{Introduction}
    In the past decade, the advent of artificial intelligence (AI) has caused major advancements in speech recognition, machine translation,  and computer vision, among other fields. One particular application has been to teach AI systems to play games. In fact, AI agents have surpassed human ability in classic arcade games and Go \cite{MnihNature} \cite{Sutton}.
    
    The most common way that these agents are trained to play games is using reinforcement learning methods. Reinforcement learning is the process where an agent in a certain state $s$ chooses to take some action $a$ in a predefined environment and receives some reward $r$. As the agent takes more actions, it is able to determine, for each $s$, which action is the best to take if the agent wants to maximize its score. 
    
    We specifically focus on how reinforcement learning can be applied to newer games like Flappy Bird. In the game, a bird moves at a constant horizontal rate and falls according to in-game gravity. There are pipes with gaps that are initialized at random points. The player must navigate the bird through the pipes by tapping on the bird to make the bird jump vertically. After overcoming each obstacle, the player gains 1 in-game point. The game ends when the bird collides with a pipe or hits the ground. The player's goal is to maximize the total score.

    In this project, we apply various AI algorithms to train an agent to play Flappy Bird. Specifically, the approaches we use are SARSA and Q-Learning. We experiment with SARSA and different variants of Q-Learning: a tabular approach, Q-value approximation via linear regression, and approximation via a neural network. In addition, we implement several modifications including $\epsilon$-greedy policy, discretization and backward updates. After comparing the Q-Learning and SARSA agents, we find that Q-Learning agents perform the best with regular scores of 1400+ and a maximum score of 2069. SARSA agents decisively outperform the baseline with a maximum score of 832.

\section{Related Work}

    We have seen successful attempts at training agents to play Atari games using reinforcement learning. Mnih et al. tried this approach on seven Atari 2600 games and found that agents trained using Q-Learning outperformed human experts on three Atari games \cite{Mnih}. Instead of engineering features from scratch to represent state $s$ and using those features to estimate $Q(s,a)$, Mnih et al. used a convolutional neural network to learn $Q(s,a)$ instead. 
    
    There have been attempts to play Flappy Bird using Q-Learning with convolutional neural networks as function approximators as well. A team tried using a similar approach to Mnih et al. by passing an image of the screen into a convolutional neural network \cite{MnihNature}. We were inspired by this project and Mnih et al. to experiment with convolutional neural networks to estimate $Q$ values. Our approach differs from these two resources in network architecture and hyperparameter choices, however.
    
    Flappy Bird was a natural choice of game, because of work done by Mnih et al. and the novelty of teaching AI agents to play newer games. While we thought that using Q-Learning with convolutional neural networks to play Flappy Bird was an interesting application, we also wanted to see how it compared to other variants of Q-Learning as well. Therefore, we took the base Q-Learning algorithm and modified it in different ways, such as adjusting our exploration probability $\epsilon$, discretizing our screen into grids of varying sizes, and using different function approximators to estimate $Q(s,a)$. We find that simpler methods like Q-Learning and SARSA outperform more complex deep Q-Learning approaches.
    
\section{Infrastructure}
We decided to use OpenAI Gym, which is an environment for training reinforcement learning agents \cite{openaigym}\cite{pygame}. OpenAI Gym provided an emulator for Flappy Bird and a useful representation of game states, namely the positions of the nearest two pipes, and the y-position of the bird. There were also images of the screen that were accessible from using the emulator. We slightly modified the state in the ways noted in section 5 and implemented a reward function.

\section{Implementation}
    Our implementation for the program was structured into the following modularized files.
    
    \begin{itemize}
        \item \texttt{main.py}: The main program that processes command line arguments and runs different agents.
        \item \texttt{FlappyBirdGame.py}: The game environment for Flappy Bird.
        \item \texttt{TemplateAgent.py}: A template for different agents.
        \item \texttt{BaselineAgent.py}: Baseline agent.
        \item \texttt{SARSAAgent.py}: SARSA agent.
        \item \texttt{QLearningAgent.py}: Q-Learning agent.
        \item \texttt{FuncApproxLRAgent.py}: Function approximation agent with linear regression.
        \item \texttt{FuncApproxDNNAgent.py}: Function approximation agent with a feed forward neural network.
        \item \texttt{FuncApproxCNNAgent.py}: Function approximation agent with a convolutional neural network.
        \item \texttt{utils.py}: A file containing some helper functions.
    \end{itemize}
    
    Link to our project: \url{https://github.com/taivu1998/FlapAI-Bird} 
    
\section{Model}
    \subsection{States}
    We came up with two formulations of state. The second formulation in Table \ref{table:stateRep2} is only used in the Q-Learning agents that use a convolutional neural network as a function approximator. The first formulation of state in Table \ref{table:stateRep1} is used in the remaining SARSA and Q-Learning agents. 
    
    Note that the $80 \times 80 \times 1$ input image has been convolved into a $17 \times 17 \times 32$ output volume and flattened. The remaining features $act$ and $y_{vel}$ are then appended to this flattened vector. Then, this whole vector is passed into the fully connected layer of the network.

    \begin{table}[h]
    \centering
    \begin{tabular}{@{}ll@{}}
    \toprule
    Feature  & Description \\ 
    \midrule
    $x_{diff}$   & The horizontal distance between the bird and the next pipe.  \\
    $y_{diff}$ & The vertical distance between the bird and the bottom pipe.\\                                                        
    $y_{vel}$ & The in-game velocity of the bird. \\
    
    \bottomrule
    
    \end{tabular}
    \\
    \caption{State representation for SARSA, Q-Learning and \\ function approximation using linear model and FFNN}
    \label{table:stateRep1}
    \end{table}

    \begin{table}[h]
    \centering
    \begin{tabular}{@{}ll@{}}
    \toprule
    Feature  & Description \\ 
    \midrule
    $image$    & A 80 $\times$ 80 grayscale image of the screen.   \\       
    $y_{vel}$ & The in-game velocity of the bird. \\
    
    \bottomrule
    
    \end{tabular}
    \\
    \caption{State representation for function approximation using CNN}
    \label{table:stateRep2}
    \end{table}
    
    \begin{figure}[!h]
    \centering
    \begin{minipage}[b]{0.4\textwidth}
    \includegraphics[width=\textwidth]{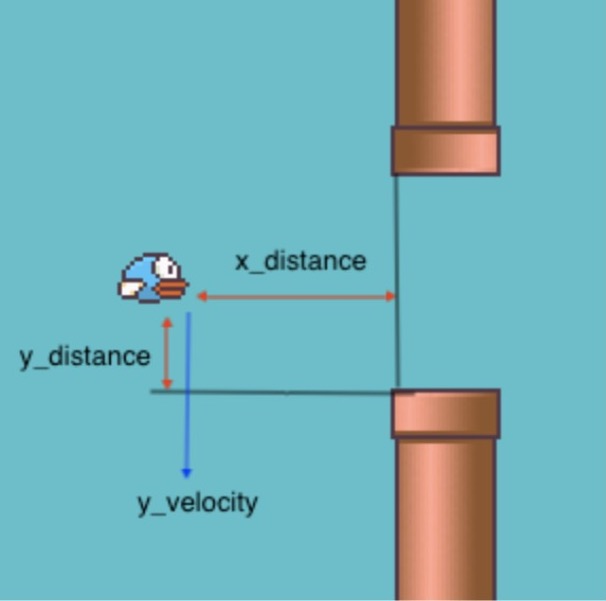}
    \caption{A graphic illustration of the first state representation}
    \end{minipage}
    \label{fig:stateRep}
    \end{figure}
    
    \subsection{Actions}
    We represented the an action $a$ using $a \in \{0, 1\}$. In particular, $0$ represents "flap", and $1$ represents "do not flap", as predefined by the game environment we used.
    
    \subsection{Rewards}
    OpenAI Gym steps through frames of the game and sets a terminal flag if the agent hits a pipe in the current frame. We decided to formulate rewards by giving the agent a positive reward of +5 every time it passes a pipe and -1000 for hitting a pipe. This is to heavily discourage the bird from making bad decisions and dying. In addition, we gave the bird a smaller reward of +0.5 for surviving in between frames. This is to incentivize the bird to stay alive even if it has not passed a pipe yet. We found that without this +0.5 reward, the bird essentially moves randomly in between pipes. This makes sense, as the bird does not prefer one state to another in between pipes, if it is not rewarded for staying alive. Only when the bird passes a pipe does it learn the optimal actions to get to the pipe in the first place.

\section{Methods}
\subsection{Baseline}
For our baseline algorithm, we simply applied a random policy. In particular, for each state, the agent selects action 0 with some probability $p$ and selects action 1 with probability $1 - p$. Here, we experimented with $p = 0.5$. We believed that this was a good baseline because it was relatively simple to implement and defined a clear strategy that a better trained agent could outperform.

 \subsection{Oracle}
 Since Flappy Bird has no maximum score, we decided that the oracle should be an agent that achieves an infinite score. Therefore, any of our agents will be outperformed by the oracle.
 
\subsection{Reinforcement Learning Algorithms}
We decided to use both on-policy and off-policy, model-free reinforcement learning algorithms to teach our agent to play Flappy Bird. We chose to do so because rewards and transitions are unknown to the agent. Therefore, we used Q-Learning and SARSA. We also made modifications to Q-Learning, by adjusting $\epsilon$, discretizing the state space, using function approximation, and using backward Q-Learning. We also modified SARSA by discretizing our screen into $10 \times 10$ grids, adding backward updates, and varying $\epsilon$. 

\subsection{SARSA}
SARSA \cite{Rummery} is an on-policy reinforcement learning algorithm that estimates $\hat{Q}_{\pi}(s,a)$ for a given state-action pair $(s,a)$, where the agent is following some policy $\pi$. SARSA takes in tuples $(a, a, r, s', a')$ and updates our estimates for $\hat{Q}_{\pi}(s,a)$ as follows:
\[\hat{Q}_{\pi}(s,a) = (1 - \eta)\hat{Q}_{\pi}(s,a) + \eta (r + \gamma \hat{Q}_{\pi}(s',a'))\]

where $r$ is the reward for taking action $a$ in state $s$, $\gamma$ is some discount value (in this case $\gamma = 1$), and for some choice of $\eta$. Intuitively, SARSA creates a bootstrapped estimate $\hat{Q}_{\pi}(s,a)$ by minimizing the difference between that estimate and the "true value" of our policy $r + \gamma\hat{Q}_{\pi}(s',a')$ over many iterations. 

\subsection{Q-Learning}
Q-Learning \cite{Watkins} is an off-policy reinforcement learning algorithm. Q-Learning creates bootstrapped estimates of $\hat{Q}_{opt}(s,a)$ by minimizing the difference between the estimate and the "true value" of our policy $\displaystyle r + \gamma \max_{a'}\hat{Q}_{opt}(s',a')$. It is similar to SARSA, except that it has the agent take the optimal action (according to our current Q-value estimates) rather than an action defined by the policy $\pi$. At the end of each iteration, our Q-value estimates $\hat{Q}_{opt}(s,a)$ are updated according to the following rule:
\[\hat{Q}_{opt}(s,a) = (1 - \eta)\hat{Q}_{opt}(s,a) + \eta (r + \gamma \max_{a'}\hat{Q}_{opt}(s',a'))\]

where $r$ is the reward for taking action $a$ in state $s$, $\gamma$ is some discount value (in this case $\gamma = 1$), and for some choice of $\eta$.

\subsection{Modifications}
We modified Q-Learning and SARSA in the following ways to speed up learning and improve performance. The implementations of the base algorithms had low performance on our task.

\subsubsection{Discretization}
Initially, we treated our screen as a $288 \times 512$ matrix of pixels, where each individual pixel as a location for our agent. We quickly found out that the number of locations were too large. Therefore we chose to treat our screen as discrete grids of sizes $5 \times 5$, $10 \times 10$, $20 \times 20$, $50 \times 50$, and $100 \times 100$ to reduce the number of states our bird could occupy. 

Specifically, for a positive integer $n$ and a level of discretization $r$, we discretize $n$ using the following formula: $$\displaystyle  d(n, r) = r \left \lfloor{\frac{n}{r}}\right \rfloor $$

\subsubsection{Epsilon-Greedy}
We implemented an epsilon-greedy approach to start off because we needed some way to explore our state space in early iterations. In our implementations of epsilon-greedy Q-Learning and SARSA, we experiment with $\epsilon \in \{0, 0.1\}$. This means with probability $\epsilon$, our agent moves randomly, and with probability $1 - \epsilon$ our agent takes an action $a^*$, where $\displaystyle a^* = \max_{a} \hat{Q}(s,a)$. We have the following algorithms for Q-Learning and SARSA in this case: 

\begin{algorithm}
	\caption{Q-Learning with epsilon-greedy policy}
	\begin{algorithmic}[1]

    \State Initialize all Q-values to 0
      \For{iteration = 1, N}
          \State Initialize memory $\mathcal{M}$
          \State Set initial state $s_1$
          \State Set $t = 1$
          \While {$s_t$ is not terminal}
              \State With probability $\epsilon$ select a random action $a_t$ from $\{0, 1\}$
              \State otherwise select $a_t = \max_a(Q(s_t, a))$
              \State Perform action $a_t$ and observe next state $s_{t+1}$ and reward $r_{t}$
              \State Store observation $(s_t, a_t, r_t, s_{t+1})$ in $\mathcal{M}$
              \State Increment $t$
          \EndWhile
          \For{observation $(s_i, a_i, r_i, s_{i+1})$ in $\mathcal{M}$}
          \State Set $\displaystyle Q(s_i, a_i) = (1-\eta)Q(s_i, a_i) + \eta (r_i + \gamma \max_{a_{i + 1}}Q(s_{i+1}, a_{i+1}))$
          \EndFor
      \EndFor
	\end{algorithmic} 
\label{fig:qlearning}
\end{algorithm}

\begin{algorithm}
	\caption{SARSA with epsilon-greedy policy}
    \begin{algorithmic}[1]

    \State Initialize all Q-values to 0
      \For{iteration = 1, N}
          \State Initialize memory $\mathcal{M}$
          \State Set initial state $s_1$
          \State Set $t = 1$
          \State With probability $\epsilon$ select a random action $a_t$ from $\{0, 1\}$
          \State otherwise select $a_t = \max_a(Q(s_t, a))$
          \While {$s_t$ is not terminal}
              \State Perform action $a_t$ and observe next state $s_{t+1}$ and reward $r_{t}$
              \State With probability $\epsilon$ select a random action $a_{t + 1}$ from $\{0, 1\}$
              \State otherwise select $a_{t + 1} = \max_a(Q(s_{t + 1}, a))$
              \State Store observation $(s_t, a_t, r_t, s_{t+1}, a_{t + 1})$ in $\mathcal{M}$
              \State Increment $t$
          \EndWhile
          \For{observation $(s_i, a_i, r_i, s_{i+1}, a_{i + 1})$ in $\mathcal{M}$}
          \State Set $Q(s_i, a_i) = (1-\eta)Q(s_i, a_i) + \eta (r_i + \gamma Q(s_{i+1}, a_{i+1}))$
          \EndFor
      \EndFor
      
    \end{algorithmic} 
\label{fig:sarsa}
\end{algorithm} 

\subsubsection{Forward versus Backward Updates}
We also experimented with backward updates. The updates were performed from the first frame to the most recent frame in forward updates. Backward updates performed Q-value updates in the opposite direction; this allows more important information, specifically when the bird hits the pipe, to be considered first.

\subsection{Function Approximation}

Similar to the original version of Q-Learning, function approximation \cite{Melo} attempts to minimize the difference between the predicted value $\hat{Q}_{opt}(s,a)$ and the target value $\displaystyle r + \gamma \max_{a'}\hat{Q}_{opt}(s',a')$. In other words, the implied loss function is 
\[\mathcal{L}(s, a, s') = \left(\hat{Q}_{opt}(s,a) - (r + \gamma \max_{a'}\hat{Q}_{opt}(s',a'))\right)^2\]

However, instead of using the above update rule, function approximation estimates the Q-values $\hat{Q}_{opt}(s,a)$ with a function $h$ and a set of weights $w$, i.e.
\[\hat{Q}_{opt}(s,a) = h(s, a; w)\]

\subsubsection{Linear Regression}
We chose to estimate our Q-values by using a linear model. Specifically, we used the first formulation of state as described in section 5.1, $\phi(s,a)$, with an action appended at the end. We took weight vector $\theta \in \mathbb{R}^3$ and a bias term $b \in \mathbb{R}$ and estimated Q-values as follows:
\[\hat{Q}(s,a) = \theta^{T}\phi(s,a) + b\]

We only used linear function approximation for Q-Learning.

\subsubsection{Feed-Forward Neural Network (FFNN)}
We also chose to use a more complex estimator in the form of a feed-forward neural network. Our network consisted of a 3-neuron input layer, a 50-neuron hidden layer, 20-neuron hidden layer, and 2-neuron output layer. If we call the size of the current layer $m$, and the size of the previous layer $n$, then each layer is modeled as follows:
\[a = f(Wa_{prev} + b)\]

where $f$ is some nonlinearity (e.g. tanh, ReLU, sigmoid). In this case, we use the ReLU functions. $a_{prev} \in \mathbb{R}^n$ is the output of the previous layer, $W \in \mathbb{R}^{m \times n}$ is the weight matrix, $b \in \mathbb{R}^m$ is the bias vector, and $y \in \mathbb{R}^m$ is the output of the current layer.

Our 2-neuron output layer was set up such that we get the Q-values of the agent being in its current state $s$ and taking action $0$ or $1$. We chose the action associated with the neuron with the highest output.

\subsubsection{Convolutional Neural Network (CNN)}
We believed that the game images were also a good way to encode state, so we used our second formulation of state in section 5.1. Since we were working with image data, a convolutional neural network was a natural choice. We preprocessed the input images by removing the background, turning them into grayscale and resizing them to 80 x 80. This helped limiting the number of parameters and eliminated unnecessary features like backgrounds and colors.

We used two convolutional layers, one using sixteen $5 \times 5$ kernels with stride $2$, and another with thirty-two $5 \times 5$ kernels with stride $2$. Finally, we flattened the output of the last convolutional layer and constructed our state vector $x$, as described in section 5.1, where $x \in \mathbb{R}^{9248}$. 

We passed $x$ through fully connected layers. In this case, our nonlinearity $f$ is the ReLU function. The fully connected layers consisted of a 9248-neuron input layer and a 2-neuron output layer.

The output of the 2-neuron layer represented the Q-values of the agent being in state $s$ and taking two actions 0 and 1.

\section{Results and Discussion}

\subsection{Model Performance: SARSA and Q-Learning}
We decided to evaluate each model based on its maximum score and average score. We chose to do so because both are important in determining which algorithm is "best" for playing Flappy Bird. The maximum score represents the peak-performance for an algorithm, while the average score is a better descriptor of the performance consistency for each agent when trained for a large number of iterations.

\begin{figure}[!h]
  \centering
  \begin{minipage}[b]{0.49\textwidth}
    \includegraphics[width=\textwidth]{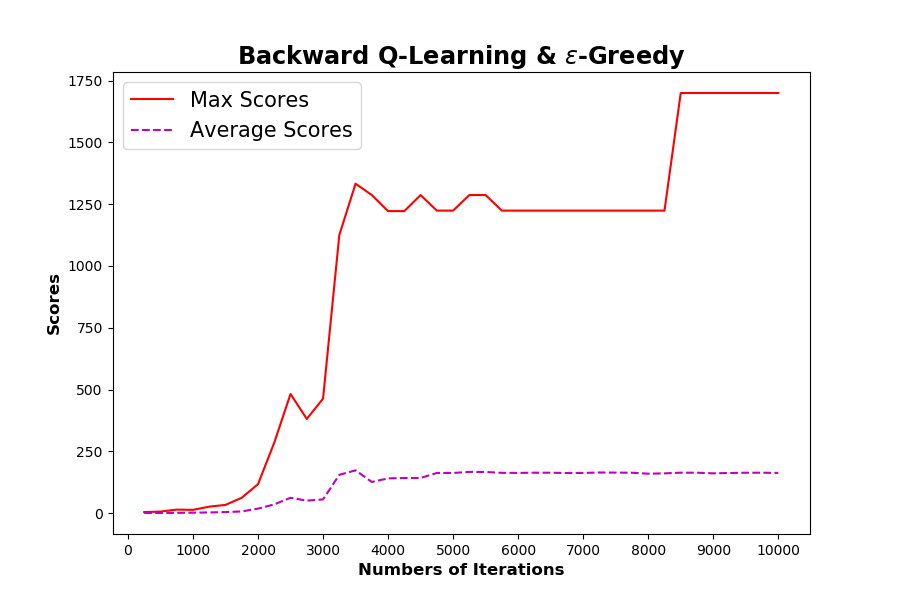}
    \caption{Training curve for Q-Learning, \\ backward updates, $\epsilon = 0.1$}
    \label{fig:backwardQLearningEpsilon}
  \end{minipage}
  \hfill
  \begin{minipage}[b]{0.49\textwidth}
    \includegraphics[width=\textwidth]{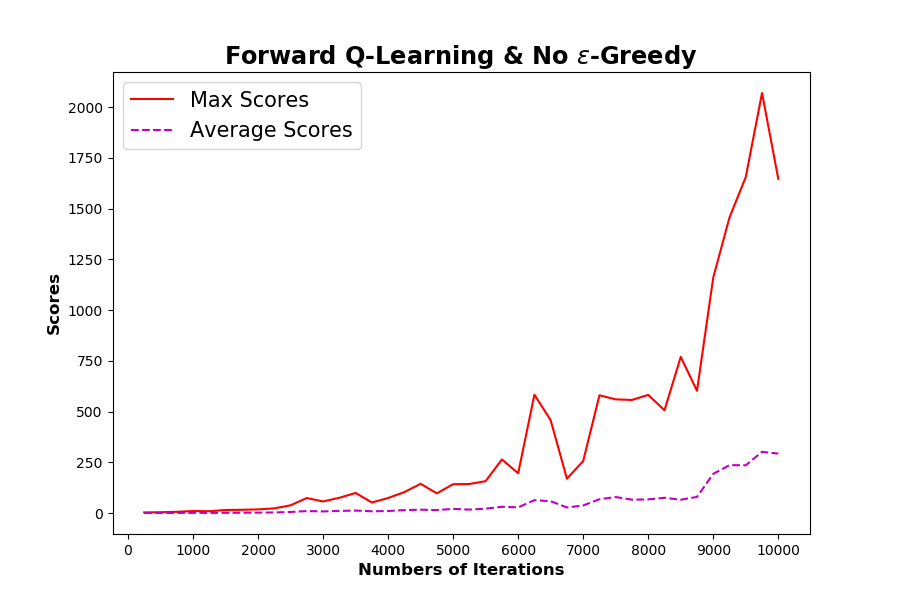}
    \caption{Training curve for Q-Learning, \\ forward updates, $\epsilon = 0$}
    \label{fig:forwardQLearningNoEpsilon}
  \end{minipage}
\end{figure}

Figures \ref{fig:backwardQLearningEpsilon} and \ref{fig:forwardQLearningNoEpsilon} show some examples of the curves that we see when training our agent for 10,000 iterations using Q-Learning. Across all variants of Q-Learning that we implemented, we see that the maximum score and average score are increasing. Especially since the average score is increasing steadily, we know that the agent is gradually learning to play the game exponentially better as we increase the number of iterations. 

\begin{table}
  \centering
  \begin{tabular}{lclcccc}
    \toprule
    \cmidrule(r){1-2}
    Algorithm     & Discretization   & Update Order  & Epsilon  & Mean & Standard Deviation & Max \\
    \midrule
    Baseline    & None &          &     & \hspace{10pt}0.130    & \hspace{10pt}0.357   & \hspace{15pt}2    \\
    Q-Learning  & 10   & Backward & 0\hspace{8pt}   & \textbf{209.298} & 216.967 & \textbf{1491} \\
    Q-Learning  & 10   & Backward & 0.1 & \textbf{159.398} & 162.553 & \textbf{1224} \\
    Q-Learning  & 10   & Forward  & 0\hspace{8pt}   & \hspace{5pt}67.210   & \hspace{5pt}69.067  & \hspace{5pt}582  \\
    Q-Learning  & 10   & Forward  & 0.1 & \hspace{5pt}63.480   & \hspace{5pt}63.346  & \hspace{5pt}448  \\
    Q-Learning  & None & Backward & 0\hspace{8pt}   & \hspace{10pt}0.507   & \hspace{10pt}0.796   & \hspace{15pt}4    \\
    Q-Learning  & None & Backward & 0.1 & \hspace{10pt}0.772   & \hspace{10pt}0.942   & \hspace{15pt}5    \\
    Q-Learning  & None & Forward  & 0\hspace{8pt}   & \hspace{10pt}0.386   & \hspace{10pt}0.644   & \hspace{15pt}4    \\
    Q-Learning  & None & Forward  & 0.1 & \hspace{10pt}0.316   & \hspace{10pt}0.593   & \hspace{15pt}3    \\
    Q-Learning  & 5    & Backward & 0\hspace{8pt}   & \hspace{5pt}36.236  & \hspace{5pt}41.438  & \hspace{5pt}438  \\
    Q-Learning  & 50   & Backward & 0\hspace{8pt}   & \hspace{5pt}80.486  & \hspace{5pt}77.624  & \hspace{5pt}622  \\
    SARSA       & 10   & Forward  & 0\hspace{8pt}   & \hspace{5pt}87.553  & \hspace{5pt}87.310   & \hspace{5pt}530  \\
    SARSA       & 10   & Forward  & 0.1 & \textbf{117.317} & 112.998 & \hspace{5pt}\textbf{811}  \\
    \bottomrule
  \end{tabular}
  \caption{Comparision across all levels of hyperparameters for SARSA and Q-Learning, 8000 iterations}
  \label{table:comparision_qlearning}
\end{table}

\subsection{SARSA versus Q-Learning}
\begin{figure}[!h]
  \centering
  \begin{minipage}[b]{0.9\textwidth}
    \includegraphics[width=\textwidth]{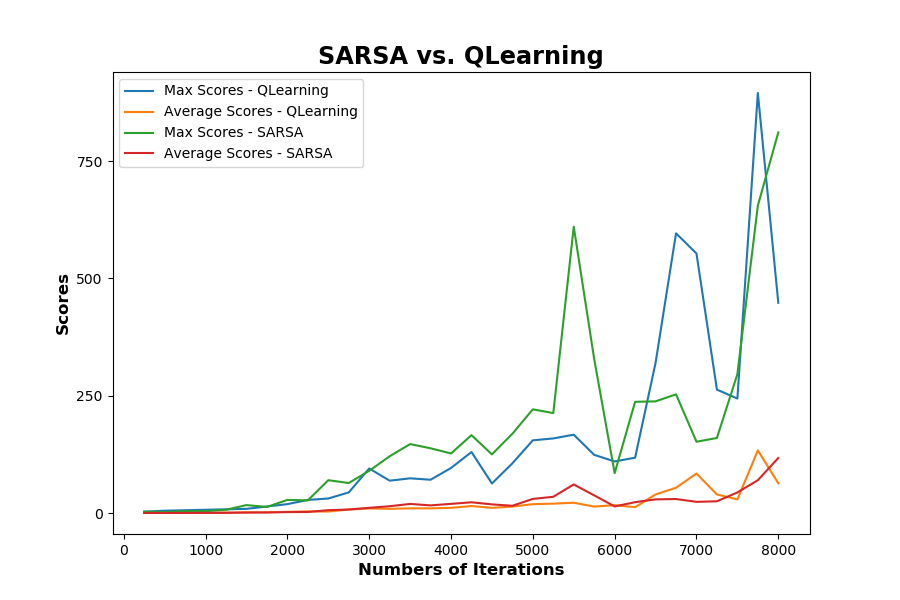}
    \caption{Training curve for Q-Learning, SARSA, forward updates, $\epsilon = 0$}
    \label{fig:SARSAvsQLearning}
  \end{minipage}
\end{figure}

We first seek to understand the difference in performance between Q-Learning and SARSA. From the plots in Figure \ref{fig:SARSAvsQLearning}, when discretization is held constant at $10 \times 10$, we see that Q-Learning has a higher maximum than SARSA earlier on. This is likely because Q-Learning is an off-policy algorithm and, thus, is able to estimate the optimal policy faster than SARSA, an on-policy algorithm can. Especially when trained for only a few thousand iterations, the agent trained by SARSA must try out both the zero and one actions at different states to find the one that yields the greatest value, rather than
following a supposed optimal policy directly, like Q-Learning can. Once we trained our SARSA agent for more iterations, though, we see that it does perform the same in terms of maximum and average score as our Q-Learning Agent. 

\subsection{Comparision of Discretization Levels}
We also chose to try out different discretizations of our screen. We experimented with $5 \times 5$, $10 \times 10$, $20 \times 20$, $50 \times 50$, and $100 \times 100$. The main trade-off that we considered was the precision of our Q-value estimate versus the reduction in state space. 

Implementing Q-Learning without discretization caused our agent to converge extremely slowly. Since our screen was $288 \times 512$, there were about $288 \times 512$ different locations the bird could be, and $16$ velocities the bird could have. Therefore, there were roughly $288 \times 512 \times 16$ states to explore. The number of states were too large, which translated to the bird only sparsely exploring the state space. For most states $s$ and action $a$, we have $Q(s,a) = 0$ because most states are unexplored. As a result, the agent would move randomly most of the time when we ran it for 8000 iterations, thereby making it no better than the baseline. It will take an unreasonably large number of iterations to explore all states. 

On the other hand, discretizing to an extreme extent, like in the $50 \times 50$ case, converges quickly but leads to a suboptimal solution. Quick convergence is the result of the agent being able to explore most of the state space. However, we make a strong, and often false, assumption that every state in a $50 \times 50$ grid has the same Q-value. Thus, we are not able to converge to the optimal solution, simply because our initial modeling assumptions were false.

\begin{figure}[!h]
  \centering
  \begin{minipage}[b]{0.9\textwidth}
    \includegraphics[width=\textwidth]{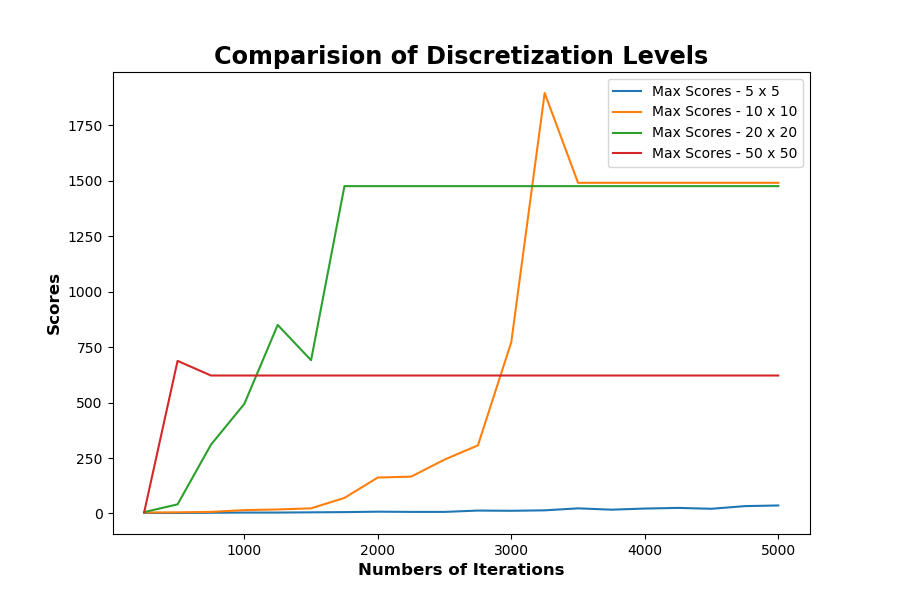}
    \caption{Maximum score achieved for Q-Learning, backward updates, $\epsilon = 0$}
    \label{fig:compareDiscretize}
  \end{minipage}
\end{figure}

Out of the different discretizations, the $10 \times 10$ discretization level performs the best, with a max score of 1896. This means that slicing the screen into $10 \times 10$ grids best trades off the Q-value estimate precision and the state space size. After a certain number of iterations, we see that each line in Figure \ref{fig:compareDiscretize} plateaus. This corresponds to convergence; once the state space is completely explored, then the agent has found the optimal policy at their level of discretization. Note that, as our discretization level become smaller, the line plateaus slower, but attains a higher score. Naturally, this follows our intuition, that no discretization, which corresponds to the $1 \times 1$ discretization level will perform the best as the number of iterations increase, but will take the longest to converge.

\subsection{Comparison of $\epsilon$ greedy policies}
Our next batch of experiments explored the effect of an epsilon-greedy policy. One of the most interesting differences came from evaluating the performance of Q-Learning with $10 \times 10$ discretization with and without an epsilon-greedy approach, where $\epsilon = 0.1$ or $\epsilon = 0$. Surprisingly, we see that the model where $\epsilon = 0.1$ performed worse than when $\epsilon = 0$. This is consistent with the cliff-walking effect detailed by Sutton and Barto \cite{Sutton}, where agents trained by Q-Learning tend to choose paths that are at the "edge" of the cliff. Our pipe represents the cliff, because it is associated with a negative value. We see that the Q-Learning agents stay very close to pipes; using an epsilon-greedy approach is almost guaranteed to cause the bird to choose a random action when it is passing through a pipe. Acting randomly inside a pipe is likely to end the game. This is reflected in our results, where the averages of the two are quite similar, 225 for $\epsilon = 0$ and 154 for $\epsilon = 0.1$. However, for maximum score, we get 1896 for $\epsilon = 0$ and 1125 for $\epsilon = 0.1$, which is a significant difference. 

\begin{figure}[!h]
  \centering
  \begin{minipage}[b]{0.49\textwidth}
    \includegraphics[width=\textwidth]{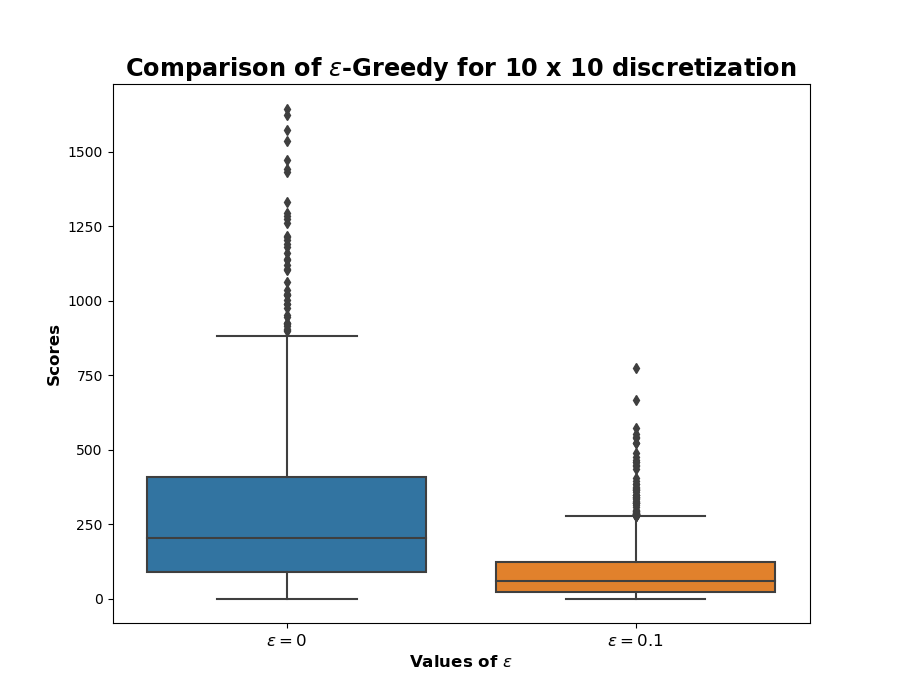}
    \caption{Epsilon comparisions for Q-Learning, \\ forward updates, $10 \times 10$ discretization}
    \label{fig:compareEpsilonQLearning}
  \end{minipage}
  \hfill
  \begin{minipage}[b]{0.49\textwidth}
    \includegraphics[width=\textwidth]{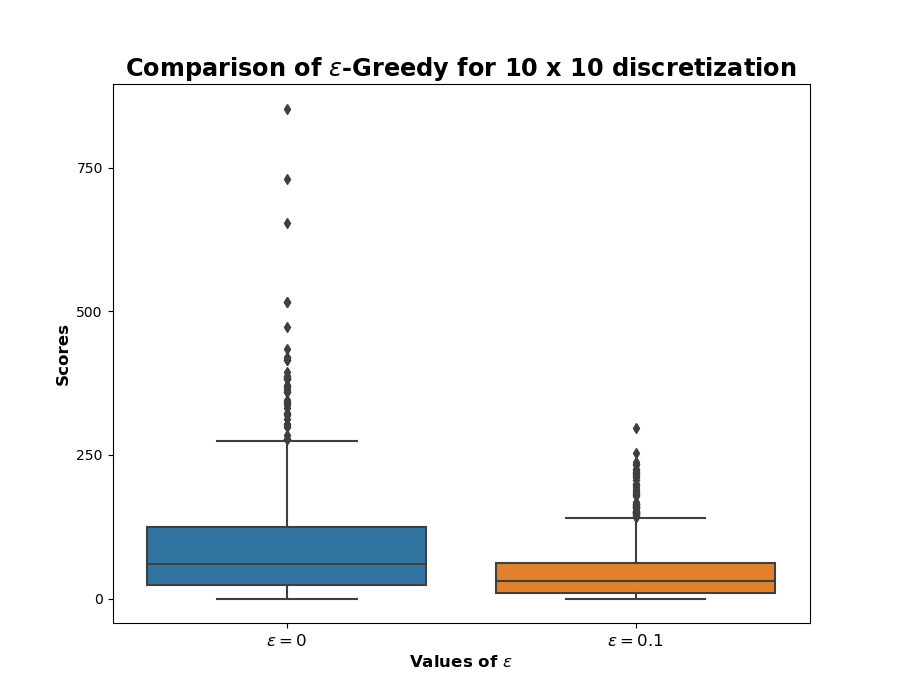}
    \caption{Epsilon comparisions for SARSA, \\ forward updates, $10 \times 10$ discretization}
    \label{fig:compareEpsilonSARSA}
  \end{minipage}
\end{figure}

In the emulator, SARSA agents try to maximize their distance from the pipes when passing in between them. This is in direct contrast to Q-Learning which remains close to one pipe. Therefore, the impact of an epsilon-greedy policy is less pronounced for SARSA. This is evident in Figures \ref{fig:compareEpsilonQLearning} and \ref{fig:compareEpsilonSARSA}, where the difference between maximum scores for SARSA of 448 versus the difference between maximum scores for Q-Learning of 987. This is because a random move is not likely to end the game if a "safe" middle path is take compared to a path close to a pipe.

In addition, both the Q-Learning and SARSA agents are able fully explore the state space, since it is discretized into $10 \times 10$ grids. Thus, there is no advantage to having an $\epsilon > 0$. Having an $\epsilon > 0$ is only useful when the state space is too large to explore in the first place.

\subsection{Comparison of Forward versus Backward Q-Learning}

\begin{figure}[]
  \centering
  \begin{minipage}[b]{0.9\textwidth}
    \includegraphics[width=\textwidth]{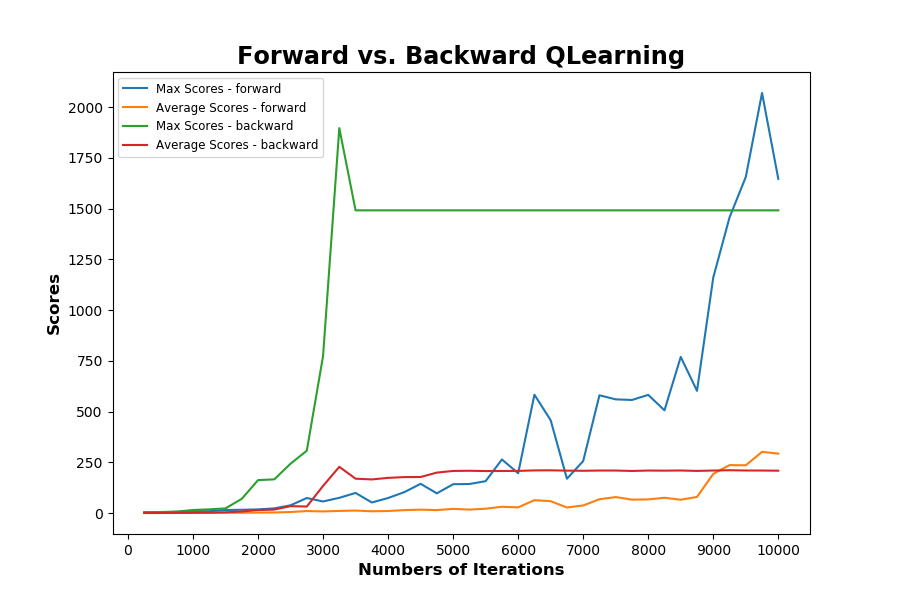}
    \caption{Comparison of forward and backward updates \\ for Q-Learning, $\epsilon = 0$, 10 $\times$ 10 discretization}
    \label{fig:forwardvsbackward}
  \end{minipage}
\end{figure}

We observe in Figure \ref{fig:forwardvsbackward} that backward Q-Learning increases the scores significantly to about 2000 after the first 3000 iterations. Then, the scores stay around 1500 for the rest of the time. Meanwhile, the scores produced by forward updates go up more gradually and stay below the figures for backward updates most of the time. However, the scores eventually reach 2000 after 10000 iterations. Hence, we can see that backward Q-Learning converges to the optimal policy much more quickly than forward Q-Learning does. Here, we define the optimal policy as the policy determined the best set of Q-values that an agent converges to after a sufficient number of iterations, given a certain level of discretization. For example, in Figure \ref{fig:forwardvsbackward}, backward Q-Learning with 10 x 10 discretization converges to the optimal policy that produces scores of around 1500 after 3500 iterations.

The reason while backward Q-Learning leads to faster convergence than forward Q-Learning does is that backward updates allow the agent to learn important information earlier. Specifically, backward Q-Learning performs updates of Q-values from the most recent and also the most important experience first, which is the frame when the bird hits the pipes. This information is then propagated through all earlier states in one iteration. Thus, the bird learns the bad states that leads to hitting the pipes more quickly and effectively, so it is able to avoid the pipes after a small number of iterations.

For example, consider a simple iteration with the sequence $s_1, a_1, r_1, s_2, a_2, r_2, s_3$, where $s_3$ is a terminal state where the bird hits the pipes. If we perform forward Q-Learning, $Q(s_1, a_1)$ will get updated first based on $s_2$ and it receives no information about the pipes at $s_3$. However, if we perform backward Q-Learning, $Q(s_2, a_2)$ will get updated first based on $s_3$ and it will change significantly because $r_2 = -1000$. Then, $Q(s_1, a_1)$ get updated based on $s_2$, so $s_1$ now gets notified about the pipes at $s_3$ via $s_2$.

\subsection{Model Performance: Function Approximation}
All our attempts at function approximation performed comparable to our baseline, which is noted in Figure 10. We believe that this is due to various shortcomings in our formulation of features and neural networks as described below. 


\begin{table}
  \centering
  \begin{tabular}{lclcccc}
    \toprule
    \cmidrule(r){1-2}
    Algorithm  & Discretization   & Update Order  & Epsilon  & Mean & Standard Deviation & Max \\
    \midrule
    Baseline          &      &          & 0.1 & 0.130  & 0.357 & 2 \\
    Linear Regression & None & Backward & 0.1 & 0.210  & 0.415 & 2 \\
    FFNN              & None & Backward & 0.1 & 0.326 & 0.437 & 3 \\
    CNN               & None & Backward & 0.1 & 0.280  & 0.372 & 3 \\
    \bottomrule
  \end{tabular}
  \caption{Comparison accross all Q-Learning agents with function approximation, backward updates, $\epsilon = 0.1$}
  \label{table:comparision_funcapprox}
\end{table}

\subsubsection{Linear Regression}
For linear regression, a line was likely too simple to capture the underlying complexities around the value of a certain state. Intutively, this makes sense; for example, for our $x_{diff}$ feature as described in section 5.1, there is not a clear linear relationship between the reward an agent receives and the distance it is from the pipe. In other words, moving closer to a pipe does not increase the expected reward of an agent by a fixed amount. Therefore, we needed a more complicated model to estimate our Q-values.

\subsubsection{Feed-Forward Neural Network (FFNN)}

After we saw that our linear regression failed to outperform the baseline, we decided to switch to a model that could model a highly nonlinear relationship between the states and the resultant Q-values. The FFNN had a similar issue as linear regression; while the model itself has higher variance, our inputs were still too simple. Essentially, we believed that only having $x_{diff}$, $y_{diff}$, and $y_{vel}$ were not enough to predict the value of state $s$. 

\subsection{Convolutional Neural Network (CNN)}

We wanted to add more relevant features than our initial three in section 5.1. Therefore, we created a second formulation of state, also described in 5.1. However, this formulation of state did not seem to make a difference, most likely because we trained it for too few iterations. Looking up related resources, we found that a deep learning approach to Flappy Bird requires learning over hundreds of thousands of iterations \cite{Lau}. We trained our network for 8000 iterations due to limited computing resources. This result was expected; no discretization was applied to the grid beforehand and, as a result, the CNN must learn the Q-values for all states in a large state space. Thus, it will take many iterations for the policy learned by the CNN to be optimal.

\section{Conclusion}

In this paper, we implement and compare the performance of agents trained by modified variants of the SARSA and Q-Learning algorithms. We notice that discretization techniques helped our agent converge more quickly within 8000 iterations and achieved a reasonably high score of 2069. We also find that, with discretization, epsilon-greedy policies generally perform worse for Q-Learning on average. Finally, we find that updating our Q-values backwards temporally helps our Q-values converge their true values faster, as well. For function approximation, we go through a natural progression of models, starting off with a simple linear model and progressing to a CNN. However, we find none of them outperformed the baseline. 

Altogether, all of our Q-Learning and SARSA agents outperform the baseline. It was only the Q-Learning with function approximation agents that either performed at or slightly above the baseline. This was largely due to poor feature engineering or lack of training. In the end, our Q-Learning agent with $10 \times 10$ discretization, backward updates, and $\epsilon = 0$ performed the best.

In the future, training the CNN for hundreds of thousands of iterations is likely to be successful. We would also want to adjust hyperparameters and experiment with more levels of discretization to see if an agent trained with smaller levels of discretization can outperform our current top model. In addition, we would like to try adding polynomial features in our linear regression and FFNN models to have them represent more complex relationships in our data.

\nocite{*}
\newpage

\bibliographystyle{ieeetr}
\bibliography{main}

\end{document}